\RequirePackage{silence}
\documentclass[runningheads]{llncs}
\usepackage{amsmath}
\usepackage{amssymb}

\usepackage[T1]{fontenc}
\usepackage{graphicx}
\usepackage{color}
\usepackage{hyperref}
\hypersetup{colorlinks=true, urlcolor=blue, linkcolor=black}

\usepackage{booktabs}
\usepackage{multirow}
\usepackage{xcolor}

\begin{document}

\title{Confidence is Not Reliability: Rethinking MC Dropout in Brain Tumour Segmentation}
\titlerunning{Confidence is Not Reliability}
\author{Xin Ci Wong \inst{1,2} \orcidID{0000-0002-1036-8023} \and Duygu Sarikaya \inst{2} \orcidID{0000-0002-2083-4999} \and Kieran Zucker \inst{3} \orcidID{0000-0003-4385-3153} \and Marc de Kamps \inst{2} \orcidID{0000-0001-7162-4425} \and Nishant Ravikumar \inst{2} \orcidID{0000-0003-0134-107X}}
\authorrunning{XC Wong et al.}
\institute{Centre for Doctoral Training in AI for Medical Diagnosis and Care, School of Computing, University of Leeds \and School of Computer Science, University of Leeds \and Leeds Cancer Centre, St James’s University Hospital, Leeds, UK
\email{scxcw@leeds.ac.uk}\\
} 
\maketitle              % typeset the header of the contribution

\begin{abstract}

Glioma segmentation in multiparametric MRI is a critical component of treatment planning. A segmentation model that fails silent\-ly on treatment-critical sub-regions represents a patient safety risk that overlap-based metrics such as Dice scores cannot expose. We ask whether voxel-level uncertainty estimation via Monte Carlo (MC) Dropout can reliably identify segmentation errors in clinically critical sub-regions, and whether calibration failure modes are detectable from standard reporting metrics alone. In an empirical two-model case study on 126 BraTS21 patients, we evaluate a high-performance pretrained SegResNet and a locally trained UNet with residual units (UNet-Res). MC dropout preserved segmentation accuracy ($|\Delta \text{Dice}|$ $<0.01$) while achieving strong uncertainty-error alignment (AUROC for entropy (H) $\approx$0.97), indicating uncertainty correctly ranks erroneous voxels above correct ones. Entropy-based patient stratification identified a high-uncertainty subgroup with substantially lower segmentation performance (median whole-tumour Dice $0.835$ \textit{vs.} $0.925$), supporting uncertainty as a practical triage signal. However, global alignment can mask important region-specific differences. Despite similar AUROC, UNet-Res exhibited near-zero enhancing tumour entropy ($0.054$) and Expected Calibration Error (ECE) of $0.915$, with a Dice of only $0.714$, indicating severely miscalibrated confidence on the most clinically critical sub-region, a failure mode invisible to standard Dice and AUROC reporting. These findings demonstrate that strong uncertainty-error alignment is necessary but insufficient for clinical safety: sub-region-specific calibration assessment must accompany AUROC evaluation when selecting models for clinical deployment.

\keywords{Brain tumour segmentation  \and Glioma MRI \and Uncertainty quantification \and Monte Carlo dropout \and Epistemic uncertainty \and Triage signal}
\end{abstract}
\section{Introduction}
Glioblastoma (GBM), which accounts for half of the adult glioma cases, is one of the deadliest yet common primary malignant tumours. The extent-of-resection (EOR) depends on accurate delineation of tumour sub-regions: whole tumour (WT), tumour core (TC), contrast-enhancing tumour (ET) from multiparametric MRI, and residual tumour at the margin can proliferate soon after surgery, placing high clinical stakes on the accuracy of each sub-region delineation \cite{Singh2025-oh}. 

Deep learning models now approach expert-level performance on benchmark datasets; however, a fundamental barrier to clinical deployment persists: deterministic predictions produce a single point-estimate per voxel with no associated confidence signal \cite{Gao2025-hn}. This is clinically consequential for two reasons. First, errors are spatially heterogeneous, and mis-labelling infiltrative tumour margins affects the EOR, while ET errors propagate into treatment response assessment. Second, without a confidence signal, models that fail on critical sub-regions do so silently, with no mechanism to trigger radiologist review before error propagate into clinical decisions. In glioma segmentation this limitation is acute: ET occupies a small fraction of total tumour voxels, and its miscalibration would be invisible to any metric computed without sub-region stratification. 

Uncertainty quantification (UQ) addresses this by providing voxel-wise confidence maps that flag regions requiring re-evaluation, without modifying acquisition protocols or requiring re-annotation. Among inference-time UQ methods, Monte Carlo (MC) Dropout \cite{Gal2016dropout} has emerged as a practical approach: by enabling dropout at inference and averaging over multiple stochastic forward passes, it approximates the posterior distribution over model weights, enabling estimation of both total predictive uncertainty and its epistemic component.

Despite increasing interest in uncertainty estimation for medical image segmentation, its evaluation is often limited to global metrics or qualitative inspection \cite{mehtaQUBraTSMICCAIBraTS2022}. These approaches may overlook clinically significant failure modes, particularly overconfident errors (low entropy, high error) in treatment-critical sub-regions such as ET \cite{Aliferis2024}. Furthermore, the behaviour of uncertainty estimates across models of differing segmentation quality has not been systematically examined. This raises two important questions: can voxel-level uncertainty estimation via MC Dropout reliably identify segmentation errors in clinically critical glioma sub-regions, and how does uncertainty behaviour vary across segmentation models of different quality and calibration? 

We address these questions through a two-model case study empirically evaluating MC Dropout on a high-performing pretrained SegResNet \cite{Myronenko2019} and a locally trained UNet-Res \cite{cardoso2022monaiopensourceframeworkdeep}, assessed on 126 patients from the BraTS21 dataset \cite{baid2021rsnaasnrmiccai}. We evaluate uncertainty quality at both voxel and patient level, and characterise model-dependent calibration behaviour across tumour sub-regions. This work makes the following contributions:
\begin{enumerate}
\item \textbf{AUROC $\neq
$ safety.} We demonstrate that strong uncertainty–error alignment (AUROC $\approx
$ 0.97) can coexist with clinically unsafe overconfidence, showing that AUROC alone is insufficient to certify model safety for sub-region-specific clinical deployment.
\item \textbf{Sub-region calibration failure invisible to standard metrics.} UNet-Res ET exhibited a calibration alert: ECE $=0.915$ with a flat reliability curve at observed fraction positive $\approx 0.40$ across all predicted probability bins, indicating that predicted probabilities carry no discriminative information about true ET voxel status, revealing a failure mode invisible to Dice and AUROC reporting alone.
\item \textbf{Entropy-based patient triage.} Entropy-based patient stratification identifies a high-uncertainty subgroup and translates voxel-level uncertainty into a deployment-ready patient-level triage signal without requiring ground-truth labels at inference time.
\end{enumerate}

\section{Related Work}

\subsection{Uncertainty quantification in medical image segmentation.}
Nair et al.\cite{Nair2020-ge} demonstrated that MC dropout can identify lesion-level segmentation failures in multiple sclerosis, providing a template for uncertainty-guided review workflows. Jungo et al.\cite{Jungo2020-rc} provided systematic analysis of uncertainty estimation quality for brain tumour segmentation, demonstrating that while voxel-wise uncertainty could be heavily miscalibrated for direct error localisation, it becomes a reliable predictor of overall segmentation failure when aggregated at the subject level, though this quality remains highly dependent on the size of the training dataset. Czolbe et al.\cite{czolbe2021uncertainty} showed that uncertainty tracks inter-observer disagreement, suggesting that uncertainty-driven workflow may not outperform random selection on genuinely ambiguous cases. Our work examines the complementary failure mode: whether uncertainty signals remain informative when model errors occur in regions of different level of ambiguity, as seen in UNet's overconfident ET predictions.

\subsection{Calibration and clinical reliability.}

Mehrtash et al. \cite{mehrtashConfidenceCalibrationPredictive2020} use expected calibration error (ECE) and reliability diagrams as standard tools for assessing whether predicted probabilities reflect empirical positive rates in medical image segmentation. This is a calibrating requirement to AUROC that captures absolute probability validity rather than ranking quality alone. Zeevi et al. \cite{zeevi2025spatiallyawareevaluationsegmentationuncertainty} subsequently showed that global calibration metrics are insufficient in dense prediction settings with severe class imbalance, where spatially heterogeneous errors can obscure sub-region-specific failure modes entirely. Alternative approaches to uncertainty estimation, including deep ensembles \cite{dwaracherla2022ensemblesuncertaintyestimationbenefits} and test-time augmentation \cite{Patel21Testime,Sherkatghanad25Testtime}, have also been explored, often demonstrating improved uncertainty quality at the cost of increased computational overhead. Prior work on sub-region-specific calibration analysis, particularly comparing ECE and reliability behaviour on ET across architectures of differing quality has not been reported \cite{Bonato2025-ts}.

\subsection{MC Dropout and practical consideration.} Gal and Ghahramani \cite{Gal2016dropout} established the theoretical grounding for MC Dropout as an approximation to Bayesian inference, showing that dropout at inference samples from an approximate posterior over model weights. Kendall and Gal \cite{Kendall17Uncertainty} extended this framework to decompose predictive uncertainty into aleatoric and epistemic components by simultaneously predicting data-dependent observation noise alongside MC Dropout sampling. An important practical extension was provided by Ledda et al. \cite{ledda2023}, who demonstrated that dropout injected post-hoc at inference, constitutes a competitive alternative to embedded dropout for epistemic uncertainty quantification, with appropriate rescaling of the uncertainty measure. This justifies post-hoc injection into pretrained segmentation models. Nair et al. \cite{Nair2020-ge} trained a 3D CNN for uncertainty measures based on MC dropout, demonstrated that small lesions and lesion borders exhibit highest uncertainty. Cross-architecture comparison of calibration behaviour, especially on sub-regions has not been systematically addressed. 

\section{Material and Methods}
\subsubsection{Datasets} The BraTS21 dataset provides $1251$ pre-operative multiparametric MRI cases (T1, T1ce, T2, FLAIR) acquired across multiple institutions\footnote{The RSNA-ASNR-MICCAI Brain Tumor Segmentation (BraTS) Challenge 2021 raw data  required to reproduce the findings of this study are available to download from \url{https://www.med.upenn.edu/cbica/brats2021/\#Data2}}. All labelled cases were partitioned into train/validation/test splits of $1000/125/126$ with fixed seed $42$. Splits were disjoint at the patient level. All volumes were skull-stripped, atlas-registered, and resampled to $1$ mm isotropic resolution at source. Preprocessing comprised per-modality z-score normalisation over non-zero (brain) voxels and centre-cropping to $128^3$ voxels.

\subsubsection{Segmentation Models} The primary segmentation model was a pretrained 3D SegResNet \cite{Myronenko2019} sourced from the MONAI \cite{cardoso2022monaiopensourceframeworkdeep} model zoo bundle \texttt{brats\_mri\_seg\-mentation v0.5.4}\footnote{The reported Dice score from MONAI official site on validation set for BraTS18: Tumor core (TC): $0.8559$; Whole tumor (WT): $0.9026$; Enhancing tumor (ET): $0.7905$; Average: $0.8518$ \url{https://github.com/Project-MONAI/model-zoo/tree/dev/models/brats_mri_segmentation}}. A custom 3D UNet with residual units (UNet-Res) \cite{cardoso2022monaiopensourceframeworkdeep} trained from scratch, provided two things: a theoretically cleaner MC Dropout baseline (dropout present during training, satisfying the Gal \& Ghahramani \cite{Gal2016dropout} assumptions), and a lower-performing regime to stress-test uncertainty calibration across model quality levels. Both models were evaluated on the same $126-$case test split, ensuring no data leakage for either model. Details implementation refer to Appendix A.

\noindent
\subsubsection{MC dropout} We activated nn.Dropout3d (p=0.2) in each model while normalisation layers remained in evaluation mode. For the pre-trained SegResNet, MC dropout was injected post-hoc at inference \cite{Gal2016dropout}. Rather than sampling from a variational posterior, each stochastic forward pass applies a random channel-level mask to intermediate activations, inducing variance across passes. The resulting uncertainty is interpreted as the model's sensitivity to feature channel ablation \textit{i.e.} regions where predictions change substantially under random masking are flagged as uncertain. For the UNet-Res, dropout was trained-in, satisfying the theoretical framework \cite{Gal2016dropout,Nair2020-ge}. Previous MC-dropout segmentation studies commonly employed $10-50$ stochastic forward passes for uncertainty estimation \cite{camarasaQuantitativeComparisonMonteCarlo2020,Nair2020-ge}. In this study, we used $20$ stochastic forward passes per patient (each sampling a different random subset of feature channels), with seeds derived as $\text{seed}_i = \text{master\_seed} + \text{patient\_idx} \times 1000 + i$ for reproducibility, allow approximated inference over a distribution of model weights \cite{Kendall17Uncertainty}. This ensemble asks: \textbf{"do different sub-networks agree on this image?"} Regions of high disagreement flag epistemic uncertainty: gaps in what the model learned from training data. No test-time augmentation was applied, isolating the MC Dropout uncertainty signal.

\noindent
\subsubsection{Uncertainty Metrics}
All uncertainty metrics were computed from the probability stack using the Bernoulli formulation per voxel per channel:

\begin{table}[ht]
    \centering
    \caption{Uncertainty type and its meaning}
    \begin{tabular}{ccc}
          Metric &  Uncertainty type & Meaning \\
\toprule
 Predictive entropy, $H_\text{pred}$ &  Total &Boundary/region needs review \\
 Expected entropy, $H_\text{exp}$  & Aleatoric & Image ambiguity \\
 Mutual information, MI & Epistemic & Model knowledge gap \\
    \end{tabular}
    \label{tab:typeuncertainty}
    
\end{table}

\begin{equation}
\bar{p} = \frac{1}{N}\sum_{i=1}^{N} p^{(i)}
\end{equation}

\noindent where $p^{(i)} \in [0,1]$ is the sigmoid output probability for a given voxel and channel at stochastic forward pass $i$, and $N=20$ is the number of passes.
 
\begin{equation}
H_\text{pred} = -\bar{p}\log\bar{p} - (1-\bar{p})\log(1-\bar{p})
\end{equation}
 
\begin{equation}
H_\text{exp} = \frac{1}{N}\sum_{i=1}^{N}
\left[-p^{(i)}\log p^{(i)} - (1-p^{(i)})\log(1-p^{(i)})\right]
\end{equation}
 
\begin{equation}
\text{MI} = H_\text{pred} - H_\text{exp} \geq 0
\end{equation}

In a standard binary classification setting, such as cancer screening, the area under the receive operating characteristic curve (AUROC) evaluates whether a model's output score can discriminate between positive (diseased) and negative (healthy) cases. In our setting, we adopt analogous formulation, where the "positive" class corresponds to erroneous voxels and the score is given by voxel-wise uncertainty, measured using entropy (H) or mutual information (MI) \cite{Nair2020-ge}. For a given uncertainty threshold, the true positive rate (TPR) represents the proportion of erroneous voxels assigned uncertainty above the threshold, while the false positive rate (FPR) represents the proportion of correctly segmented voxels exceeding the same threshold. AUROC summarises this relationship across all possible threshold and can be interpreted as the probability that a randomly selected erroneous voxel has higher uncertainty than a randomly selected correct voxel, \textit{i.e.} does uncertainty detect errors? In dense segmentation tasks, erroneous voxels typically constitute a small fraction of the total volume. While AUROC remains a useful measure of uncertainty-error ranking, it does not capture the absolute reliability of uncertainty estimates. In particular, a model can achieve high AUROC by correctly ranking most errors above correct voxels, while still assigning low uncertainty to subset of clinically critical error. This limitation motivates the inclusion of calibration metrics, such as Expected Calibration Error (ECE), to assess whether predicted confidence is aligned with empirical correctness.

ECE measures whether the predicted probabilities are trustworthy. It is the average of gaps on the reliability diagram, and was computed on foreground-relevant voxels defined as the union of predicted positive voxels ($\bar{p}>0.1$) and GT-positive voxels, using $K=20$ equal-width bins, \textit{i.e.} how well a model’s predicted probabilities match the true outcome frequencies \cite{mehrtashConfidenceCalibrationPredictive2020}. Background exclusion was required as 98.5\% of voxels have $\bar{p}<0.05$. Population ECE was computed as a voxel-count-weighted average of per-patient bin statistics. Sensitivity to the foreground
threshold was evaluated across $\bar{p} \in \{0.05, 0.10, 0.20, 0.30\}$ (Appendix B).

\begin{equation}
    ECE = \sum_{b=1}^{B} \frac{|B_b|}{N} \cdot \left| \bar{p}_b - \bar{a}_b \right|
\end{equation}

where $B_b$ is the $b$-th probability bin, $\bar{p}_b = \frac{1}{|B_b|}\sum_{i \in B_b} \hat{p}_i$ is the mean predicted probability in bin $b$, and $\bar{a}_b = \frac{1}{|B_b|}\sum_{i \in B_b} \mathbf{1}[y_i = 1]$ is the observed accuracy in bin $b$ for binary GT labels, \textit{i.e.} the  fraction of voxels in this bin that are truly positive. The $\bar{p}$ is the mean sigmoid output across 20 MC Dropout passes for a given voxel and a given channel, \textit{i.e.} the average predicted probability that this voxel belongs to certain sub-regions. 

\section{Results}
\subsection{Segmentation performance}

\begin{figure}
    \centering
    \includegraphics[width=1.0\linewidth]{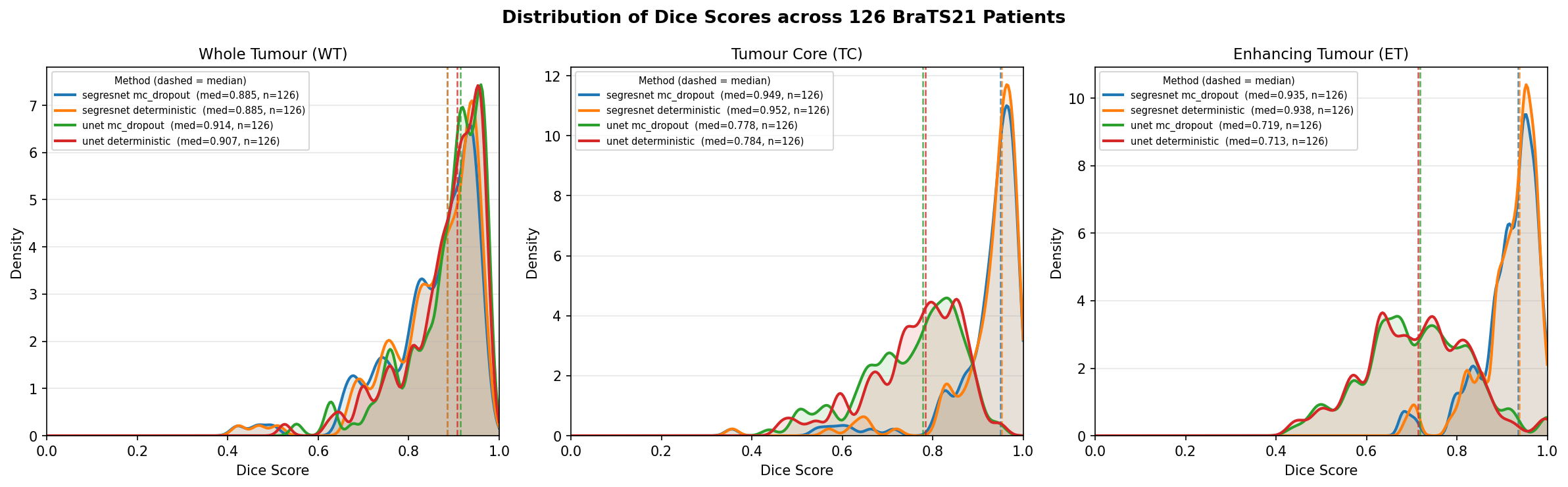}
    \caption{\textbf{Distribution of Dice scores (n=126 patients per model)}. Kernel density estimates show per-method score distributions for whole tumour(WT), tumour core (TC) and enhancing tumour (ET). Dashed vertical lines indicate per-method medians. SegResNet methods clustered tightly near the deterministic baseline on all sub-regions; UNet-Res shows a broad, bimodal distribution on TC and ET with a tail below Dice 0.6, indicating a subset of patients with poor sub-region discrimination. MC Dropout and deterministic inference are visually indistinguishable within each architecture, confirming that stochastic inference does not shift the performance distribution.}
    \label{fig:kde}
\end{figure}

SegResNet MC Dropout achieved Dice of $0.853 \pm 0.103$ (WT), $0.918 \pm 0.097$ (TC), and $0.918 \pm 0.064$ (ET). UNet-Res MC Dropout achieved higher WT Dice ($0.885 \pm 0.086$) but substantially lower TC ($0.756 \pm 0.108$) and ET ($0.714 \pm 0.117$), consistent with a trade-off between global spatial context and sub-region discrimination (Table \ref{tab:dice}). The UNet-Res Dice distributions on TC and ET are bimodal, with a secondary peak below Dice~0.6 (Fig. \ref{fig:kde}), indicating a clinically relevant failure mode in a minority of patients. Paired Wilcoxon signed-rank tests (Bonferroni-corrected) confirmed that stochastic inference introduced statistically detectable but
clinically negligible accuracy differences relative to deterministic baselines ($|\Delta| < 0.01$ across all sub-regions for both models), confirming MC Dropout can be deployed without segmentation quality
penalty (Table \ref{tab:dice}).

\begin{table}[ht]
    \centering
    \caption{Segmentation performance (Dice $\uparrow$, mean $\pm$ SD) on BraTS2021 (126 patients). Paired Wilcoxon signed-rank tests compare $\text{MC Dropout}^\ddagger$ with $\text{deterministic}^\dagger$ inference (Bonferroni-corrected). While several differences are statistically significant, effect sizes are small ($|\Delta| < 0.01$), indicating negligible practical impact on segmentation performance. Best results per model are highlighted in bold.}
    \label{tab:dice}
    \scriptsize
    \setlength{\tabcolsep}{5pt}
    \renewcommand{\arraystretch}{1.2}

    \begin{tabular}{lccc|ccc}
    \toprule
    \textbf{Model} &  WT & TC & ET 
& WT & TC & ET \\
   &  &  &  & \multicolumn{3}{c}{\textbf{Wilcoxon (MC \textit{vs.} Det)}} \\
    \midrule

% SegResNet
$\text{SegResNet}^{\dagger}$ 
  & \textbf{0.860 $\pm$ 0.100} & \textbf{0.922 $\pm$ 0.092} & \textbf{0.920 $\pm$ 0.062}
  & --- & --- & --- \\

$\text{SegResNet}^{\ddagger}$ 
  & 0.853 $\pm$ 0.103 & 0.918 $\pm$ 0.097 & 0.918 $\pm$ 0.064
  & *** & *** & *** \\

    \midrule

% UNet
$\text{UNet-Res}^{\dagger}$
  & 0.880 $\pm$ 0.087 & \textbf{0.762 $\pm$ 0.104} & 0.710 $\pm$ 0.115
  & --- & --- & --- \\
$\text{UNet-Res}^\ddagger$
  & \textbf{0.885 $\pm$ 0.086} & 0.756 $\pm$ 0.108 & \textbf{0.714 $\pm$ 0.117}
  & ** & *** & ns \\

    \bottomrule
    \end{tabular}

    \smallskip
    \raggedright
    \footnotesize
    $^\dagger$~=~Deterministic;$^\ddagger$~=~MC Dropout;
    \textit{WT}~=~whole tumour; \textit{TC}~=~tumour core; \textit{ET}~=~enhancing tumour; \textit{MC}~=~MC dropout; \textit{Det}~=~Deterministic model. ns = not significant; * $p<0.05$, 
** $p<0.01$, *** $p<0.001$. $\Delta$ denotes MC Dropout minus deterministic Dice.
\end{table}

\subsection{Uncertainty quality} 
Both models achieved strong entropy-error alignment at voxel level: AUROC-H was $0.977 \pm 0.028$ (SegResNet) and $0.975 \pm 0.037$ (UNet-Res); AUROC-MI was $0.981 \pm 0.020$ and $0.969 \pm 0.038$ respectively (Table \ref{tab:uncertainty}). The comparable AUROC values between architectures,
however, may mask a critical divergence in calibration quality and spatial behaviour that has direct clinical implications.

\begin{table}[t]
\centering
\caption{Uncertainty quality evaluation (mean $\pm$ SD) for MC dropout method. Entropy (Ent) is voxel-wise predictive entropy. AUROC evaluates uncertainty–error alignment. Lower entropy indicates more confident predictions; higher AUROC indicates better alignment with segmentation errors.}
\label{tab:uncertainty}
\scriptsize
\setlength{\tabcolsep}{5pt}
\renewcommand{\arraystretch}{1.2}

\begin{tabular}{llrccc cc}
\toprule
\textbf{Model}  & Ent(WT) $\downarrow$ & Ent(TC)$\downarrow$ & Ent(ET) $\downarrow$
& AUROC-H $\uparrow$ & AUROC-MI $\uparrow$ \\
\midrule

% SegResNet
SegResNet  & 0.296 $\pm$ 0.071 & 0.277 $\pm$ 0.107 & 0.336 $\pm$ 0.111
  & 0.977 $\pm$ 0.028 & 0.981 $\pm$ 0.020 \\

\midrule

% UNet
UNet-Res  &  0.113 $\pm$ 0.074 & 0.269 $\pm$ 0.128 & 0.054 $\pm$ 0.060
  & 0.975 $\pm$ 0.037 & 0.969 $\pm$ 0.038 \\

\bottomrule
\end{tabular}

\smallskip
\raggedright
\footnotesize
Ent = voxel-level Shannon entropy; AUROC-H = entropy \textit{vs.} error AUROC; AUROC-MI = mutual information \textit{vs.} error AUROC. 

\end{table}

\begin{figure}[ht]
    \centering
\includegraphics[width=1.0\linewidth]{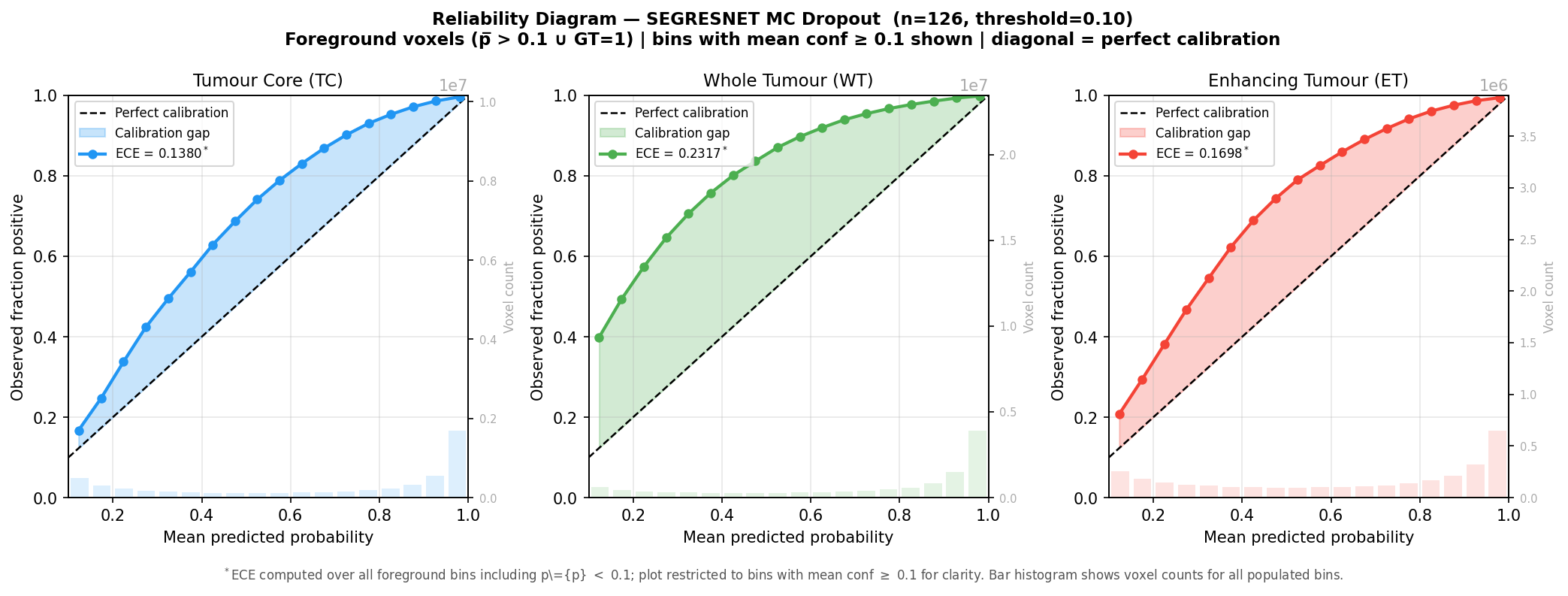}
\includegraphics[width=1.0\linewidth]{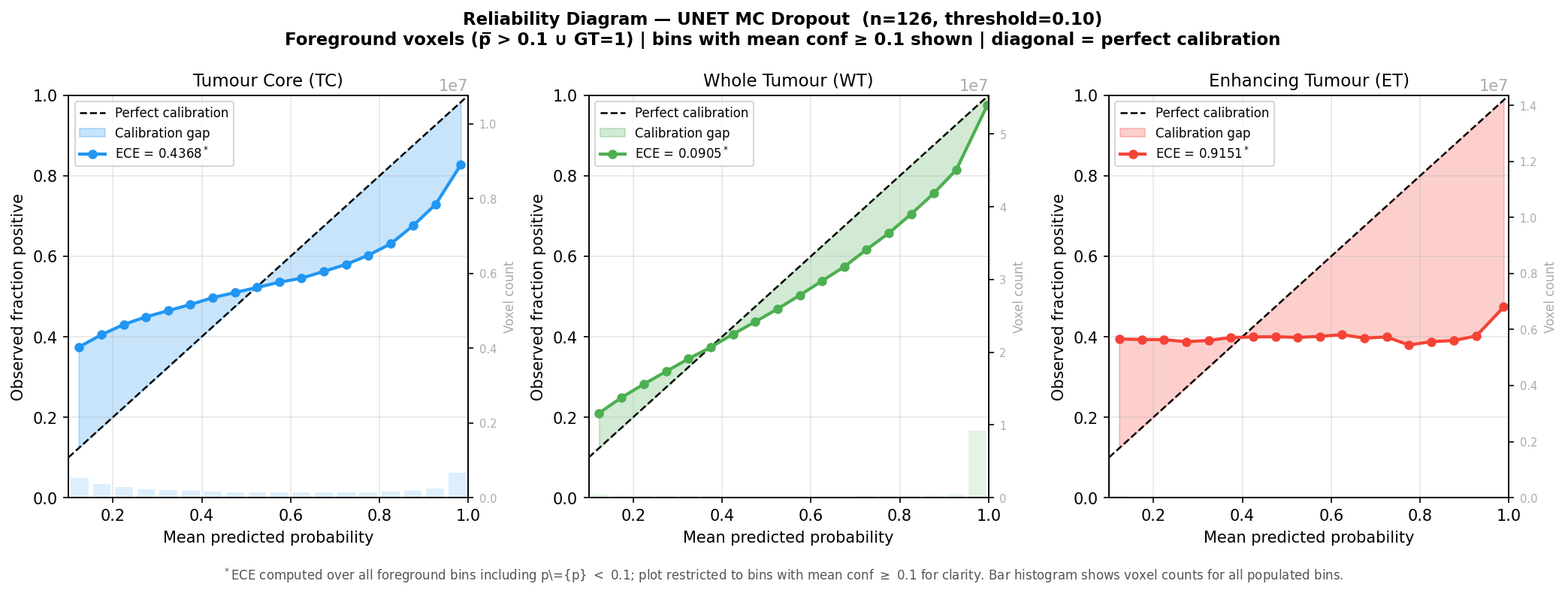}
   \caption{\textbf{Reliability diagrams computed on foreground-relevant voxels (predicted positive p > 0.1 $\cup$ GT positive, K=20 equal-width bins)}. Each point represents one bin of foreground voxels. ECE values ($*$) are computed over all foreground bins. The bar histogram on the secondary y-axis shows how many foreground voxels fall in each probability bin, revealing where the model concentrates its predictions. Curves above the diagonal indicate underconfidence across all sub-regions; UNet-Res ET (ECE = 0.915) exhibits near-constant acc $\approx$ 0.4 regardless of predicted probability, confirming the overconfident-but-wrong behaviour identified in entropy analysis. }
    \label{fig:reliability}
\end{figure}

Calibration analysis (Fig. \ref{fig:reliability}) confirmed
this divergence. SegResNet showed moderate, consistent under-confidence across all sub-regions (ECE: WT=0.232, TC=0.138, ET=0.170), with reliability curves above the diagonal throughout. UNet-Res WT was the best-calibrated result across all measurements (ECE$=0.091$). UNet-Res ET, however, showed critical miscalibration (ECE$>0.9$): the reliability diagram reveals an approximately flat curve at observed fraction positive $\approx 0.40$ regardless of predicted probability across the full $[0.1, 0.9]$ range, indicating that UNet-Res's ET probability outputs carry no discriminative information about true ET voxel status. Sensitivity analysis across foreground thresholds $\bar{p} \in \{0.05, 0.10, 0.20, 0.30\}$ confirmed the finding was threshold-independent (UNet-Res ET ECE range 0.888--0.948 versus 0.141--0.230 for SegResNet ET).

\subsection{Overconfidence on Enhancing Tumour}
\begin{figure}
    \centering
    \includegraphics[width=1.0\linewidth]{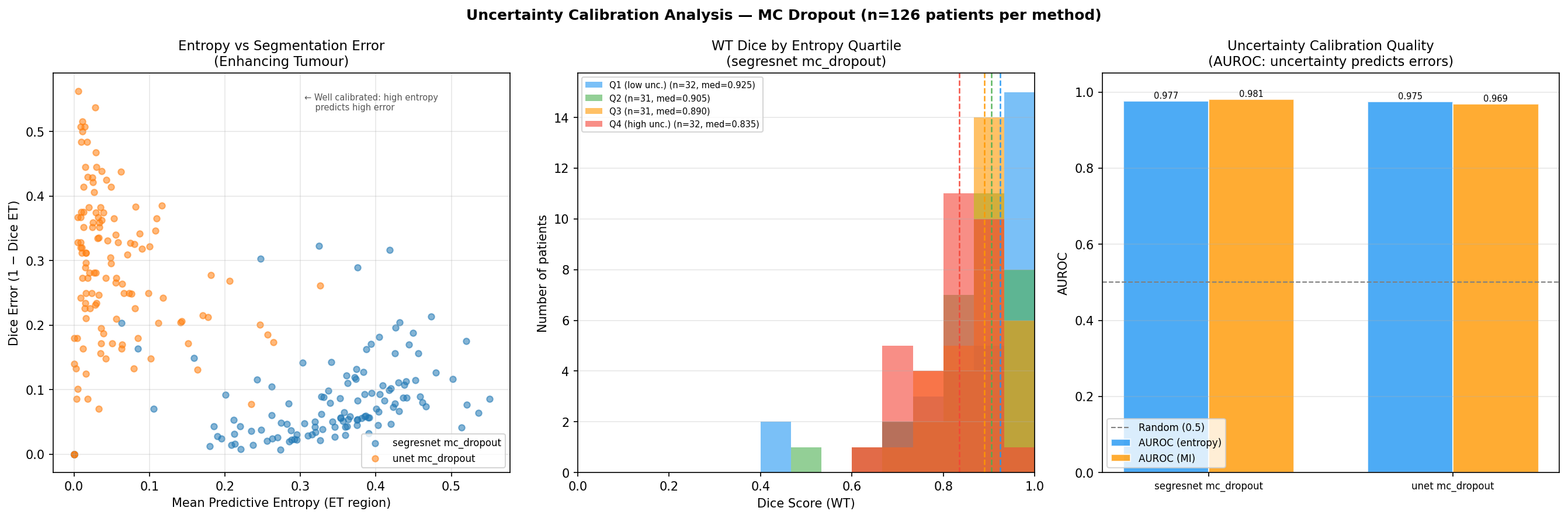}
    \caption{\textbf{Uncertainty analysis for MC Dropout (n=126 patients per model)}. Left: per-patient mean predictive entropy versus ET Dice error; SegResNet (blue) shows positive entropy-error correlation while UNet-Res (orange) clusters at near-zero entropy regardless of error magnitude, indicating systematic overconfidence on the most treatment-critical sub-region.  Centre: WT Dice stratified by SegResNet MC Dropout entropy quartile; median Dice decreases monotonically from Q1 (0.925) to Q4 (0.835), demonstrating uncertainty as a clinically usable triage signal. Right: AUROC for entropy and mutual information versus voxel-level error for both models. }
    \label{fig:uncertainty}
\end{figure}

\begin{figure}
    \centering
    \includegraphics[width=1.0\linewidth]{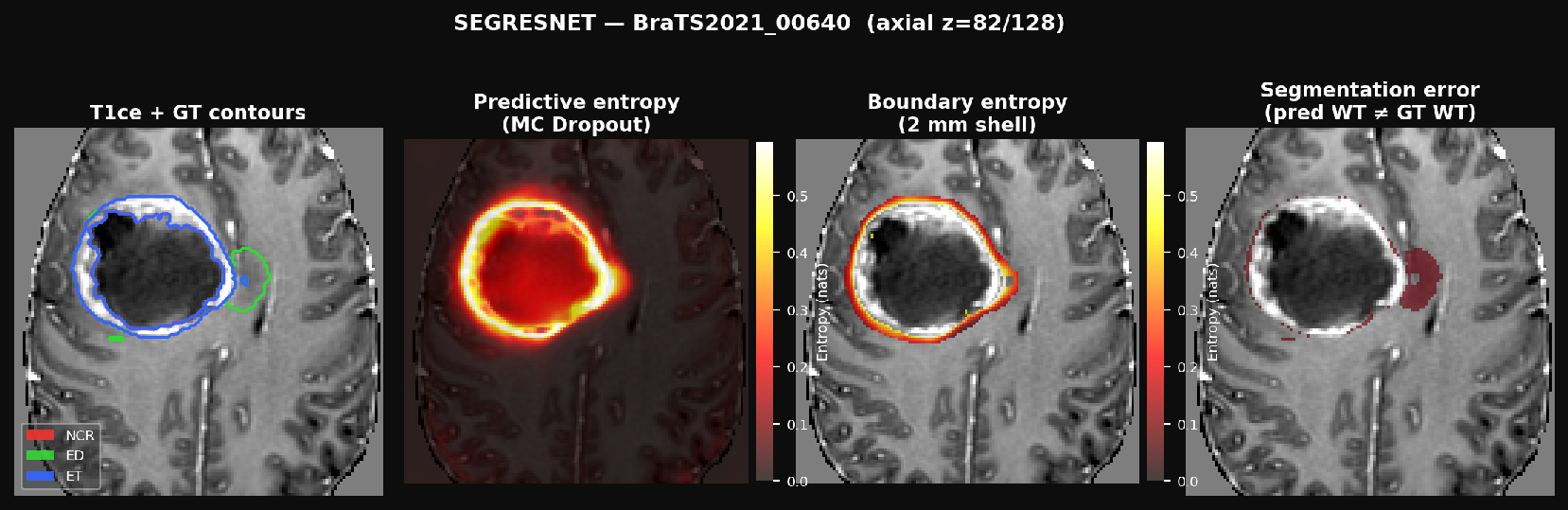}
    \includegraphics[width=1.0\linewidth]{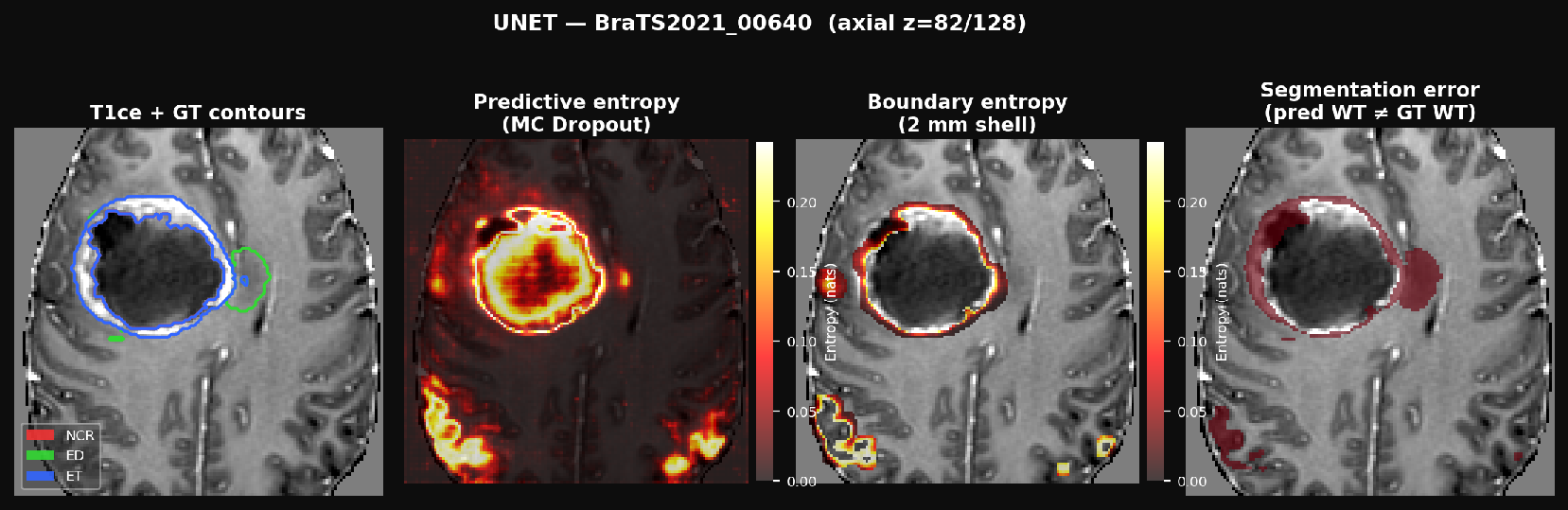}
    \caption{\textbf{Qualitative comparison of segmentation and uncertainty estimation for a representative BraTS2021 case (BraTS2021\_00640, axial z=82)}. From Left to Right: T1ce + Ground Truth (GT) contours, Entropy overlay, Boundary entropy, Segmentation error map. GT shown as contour lines not filled regions, preserves anatomy visibility underneath. Warmer colours indicate higher predictive entropy at the predicted tumour margin, identifying regions where the model's confidence in the delineation boundary is lowest. UNet-Res shows high entropy not only at the boundary, but also concentrated at the tumour core, while SegResNet's entropy is a clean ring at the tumour-brain interface. }
    \label{fig:heatmap}
    
\end{figure}

The ET entropy--error scatterplot (Fig. \ref{fig:uncertainty}, left)
reveals the central finding of this study. UNet-Res assigns near-zero
ET entropy regardless of segmentation error magnitude: patients with
greater than 40\% ET Dice error are indistinguishable in uncertainty
from well-segmented cases (UNet-Res ET entropy = $0.054 \pm 0.060$).
In contrast, SegResNet shows a positive entropy--error correlation
across a wide entropy range, providing an actionable clinical triage signal despite achieving substantially
higher ET Dice ($0.918$ \textit{vs.} $0.714$). This dissociation between
accuracy and uncertainty behaviour is the defining finding: high AUROC
with low entropy magnitude produces a signal insufficient for clinical
thresholding. Qualitative uncertainty maps (Figs. \ref{fig:heatmap}) further illustrate the spatial divergence.

\begin{figure}
    \centering
    \includegraphics[width=1.0\linewidth]{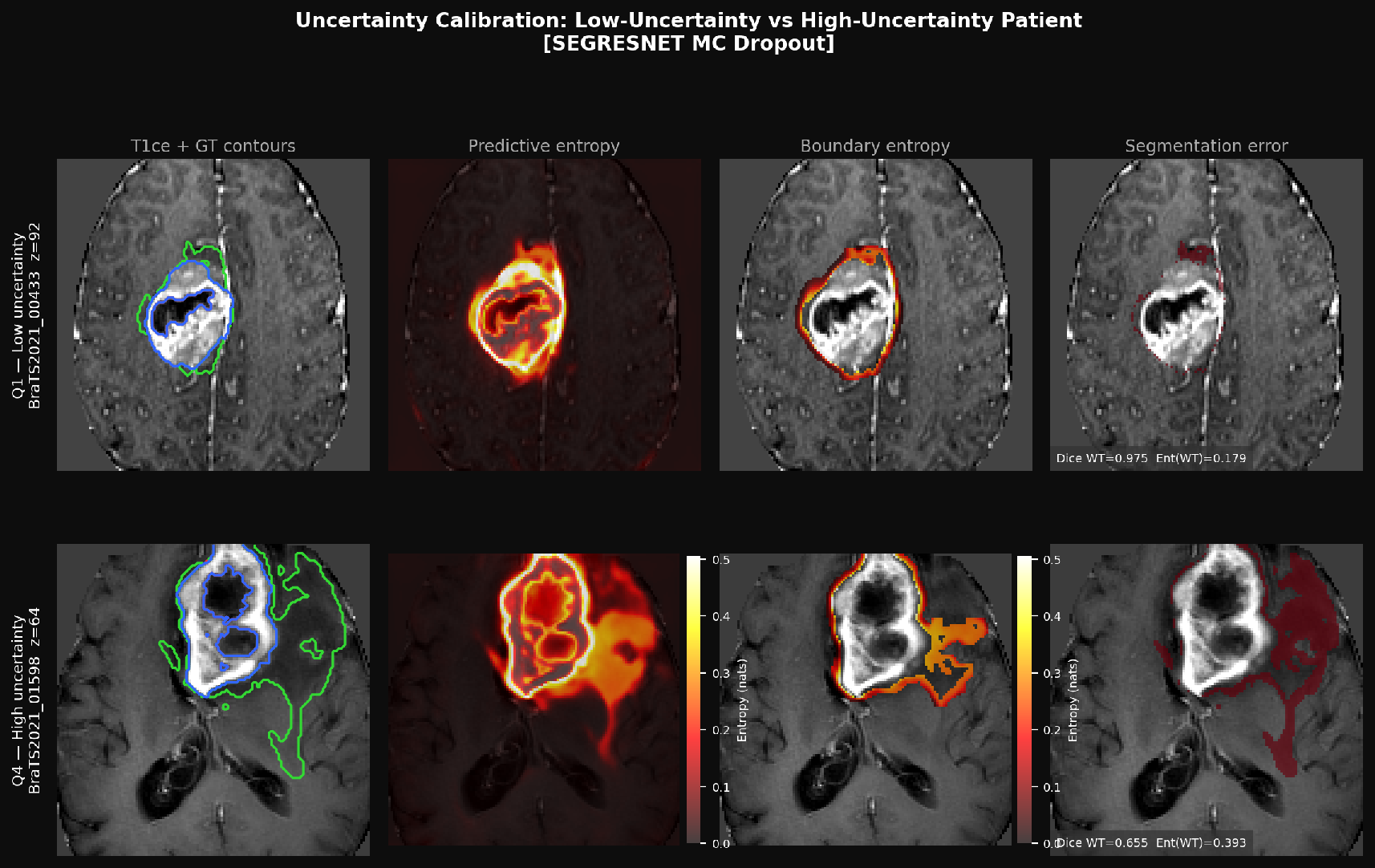}
    \includegraphics[width=1.0\linewidth]{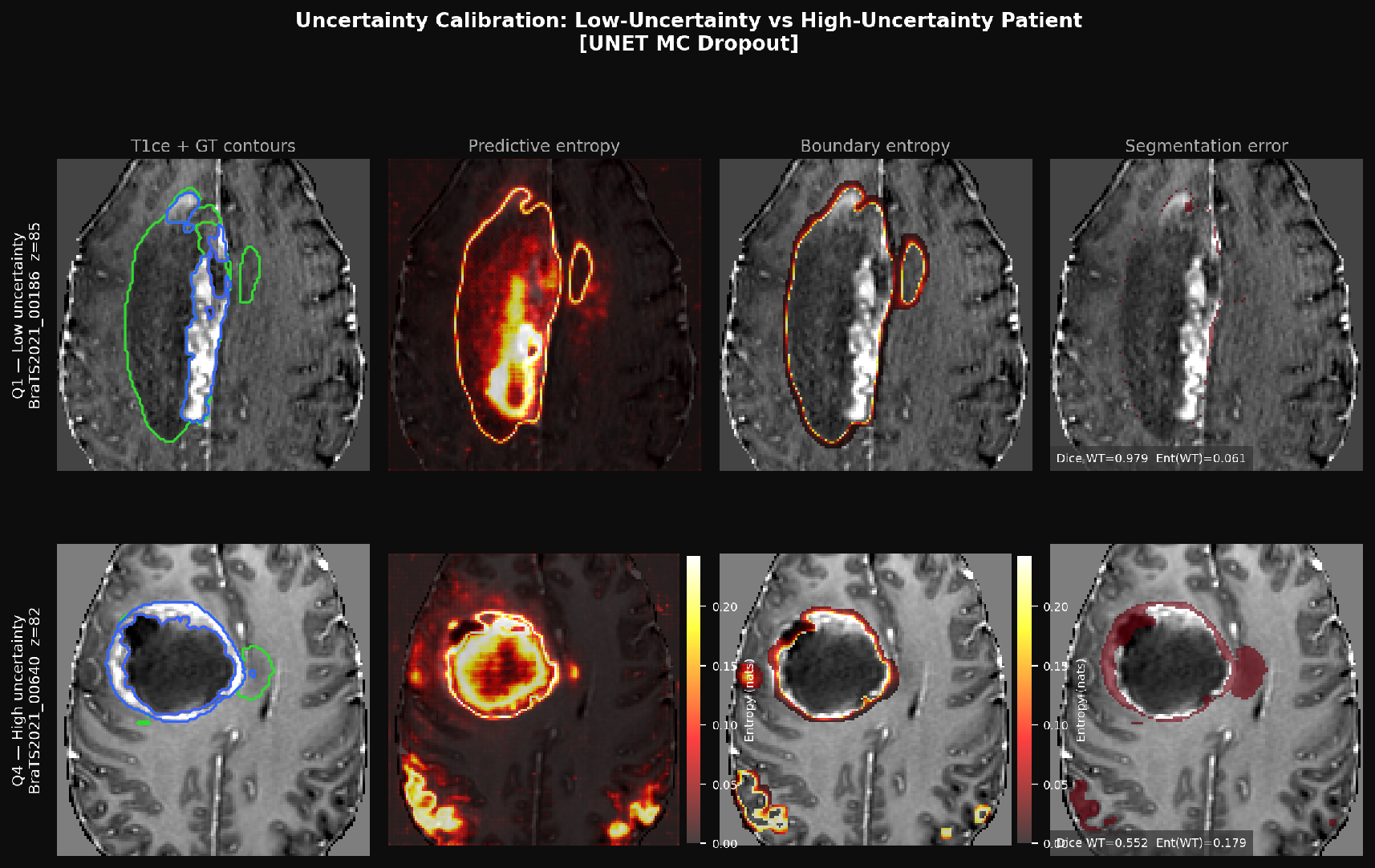}
    \caption{\textbf{Uncertainty heatmap of Low-Uncertainty \textit{vs.} High Uncertainty Patient.} For each SegResNet and UNet-Res, Q1 patient (top row): Low entropy, minimal segmentation error, clean boundary. Entropy is relatively faint, boundary entropy ring is thinner and dimmer. Q4 patient (bottom row): Entropy (0.393 \textit{vs.} 0.179) both are badly failing cases. The segmentation error shows massive dark red area. SegResNet also expresses more uncertainty even on its well-segmented case. }
    \label{fig:comparison}
    
\end{figure}

\subsection{Patient-Level Triage}

The entropy quartile analysis (Figure \ref{fig:uncertainty}, centre) demonstrates clinical utility directly: SegResNet patients in the highest uncertainty quartile had median WT Dice of 0.835 versus 0.925 in the lowest quartile, indicating that uncertainty stratification identifies a subset at substantially elevated error risk.

The calibration comparison figures (Figs. \ref{fig:comparison}) contrast representative Q1/Q4 patients per model. For SegResNet, the Q4 case presents a morphologically complex multifocal tumour (BraTS2021\_01598, Dice WT=0.655, Ent=0.393): elevated entropy is spatially co-localised with errors at lesion margins and the
inter-lesion parenchyma, correctly flagging an atypical case. For UNet-Res, the Q4 case (BraTS2021\_00640, Dice WT=0.552, Ent=0.179) shows that despite catastrophic WT failure, mean entropy increases only 3-fold relative to Q1, and the entropy remains spatially misaligned with the principal error region. The model does not adequately signal the degree of failure.

\section{Discussion and Conclusion}

\subsubsection{Clinical implications of overconfidence} The principal finding is a dissociation between segmentation accuracy and uncertainty behaviour with direct patient safety implications. UNet-Res achieves competitive WT Dice yet is systematically onverconfident on ET, the sub-region used to assess chemotherapy response and guide stereotactic biopsy targeting for RANO criteria\footnote{The RANO (Response Assessment in Neuro-Oncology) criteria are standardised guidelines used to evaluate treatment response in glioma trials, focusing on tumor measurement via MRI.} \cite{Sakata2025-zf}. A model that assigns equivalent confidence to a 40\% ET error and a good  segmentation (Deterministic WT Dice: $0.880\pm0.087$) cannot support safe semi-automated workflows, regardless of its mean Dice performance. This finding demonstrates why model selection for clinical deployment must consider uncertainty behavior on sub-region level, not only overlap-based Dice scores \cite{czolbe2021uncertainty}. The entropy quartile analysis translates uncertainty quantification into a clinically actionable finding: patients flagged as high-uncertainty by MC Dropout have demonstrably worse segmentation, enabling targeted radiologist review without requiring full manual re-segmentation of every case\cite{Nair2020-ge}. This triage function \textit{i.e.} directing attention to the cases most likely to contain errors, addresses a practical barrier to clinical deployment of automated segmentation\cite{Jungo2020-rc}. 

\subsubsection{Why AUROC alone is insufficient} Both models achieve AUROC-H $\approx 0.975$, which may appear to indicate equivalent clinical utility with good uncertainty-error alignment. However, the voxel-level class imbalance may inflate absolute values, even a weak uncertainty signal can achieve high AUROC by correctly ranking a few erroneous voxels above the large pool of correct ones \cite{czolbe2021uncertainty}. ECE analysis reveals why AUROC could be misleading. High ECE for UNet-Res ET with a flat reliability curve demonstrates that the model's ET probability outputs are informationally invalid, \textit{i.e.} the predicted probabilities do not reflect true positive rates. A model can rank well but be poorly calibrated. Therefore, ECE and entropy magnitude must be reported alongside AUROC for a complete calibration picture; AUROC alone is insufficient to certify clinical safety \cite{mehrtashConfidenceCalibrationPredictive2020}.

\subsubsection{Limitations} Evaluation is limited to a single publicly available dataset, which uses consensus labels.  The 20-pass MC Dropout approximation was not evaluated for sensitivity to pass count, and post-hoc dropout injection into SegResNet weakens direct comparison with the embedded-dropout UNet-Res. Future work should incorporate multi-site data, calibration metrics with temperature scaling or Platt scaling to reduce ECE, and prospective reader studies evaluating radiologist efficiency with uncertainty overlays.

\subsubsection{Conclusion} We evaluated MC Dropout uncertainty estimation across SegResNet and UNet-Res on 126 BraTS21 test set, demonstrating that high uncertainty-error alignment (AUROC-H $\approx 0.97$) can coexist with clinically unusable uncertainty signals. The central finding is a dissociation between ranking quality and calibration validity: despite comparable AUROC, the lower-performing architecture exhibited near-zero ET entropy (0.054), ECE of 0.915, and a flat reliability curve, indicating that its ET probability outputs carry no discriminative information, rendering the uncertainty signal uninformative on the sub-region most directly linked to treatment decisions. This failure mode was invisible to Dice and AUROC reporting alone. Furthermore, the UNet-Res was simultaneously better calibrated on WT yet critically miscalibrated on ET, demonstrating that sub-region performance and calibration can vary within the same model. Entropy-based patient stratification provided a practical triage signal, translating uncertainty into a patient-level review trigger.

\begin{credits}
\subsubsection{\ackname} We acknowledge the support of the Centre for Doctoral Training in AI for Medical Diagnosis and Care for funding this project. We extend our gratitude to the patients and doctors who contributed to datasets. The study was facilitated using the Aire High Performance Computing resources at the University of Leeds, UK. 

\subsubsection{\discintname}
KZ is a founder, director, and shareholder of Synaptome Limited. NR is co-founder and shareholder of adsilico Ltd. The remaining authors declare no competing interests relevant to the content of this article.
\end{credits}
\bibliographystyle{splncs04}
\bibliography{mybibliography}

\end{document}